\documentclass{aamas2012_modified}

\usepackage{graphicx}
\usepackage{subfigure}
\usepackage{amssymb}
\usepackage{amsmath}

\newcommand{\tuple}[1]{\ensuremath{\langle #1 \rangle }}

\begin{document}

\title{Framing Human-Robot Task Communication as a POMDP}

\numberofauthors{2}

\author{
\alignauthor
Mark P. Woodward\\
\affaddr{Harvard University}\\
\affaddr{60 Oxford St.}\\
\affaddr{Cambridge, Massachussettes 02138 USA}\\
\email{mwoodward@eecs.harvard.edu}
\alignauthor
Robert J. Wood\\
\affaddr{Harvard University}\\
\affaddr{33 Oxford St.}\\
\affaddr{Cambridge, Massachussettes 02138 USA}\\
\email{rjwood@eecs.harvard.edu}
}

\maketitle

\begin{abstract}
As general purpose robots become more capable, pre-pro-gramming of all tasks at the factory will become less practical. We would like for non-technical human owners to be able to communicate, through interaction with their robot, the details of a new task; we call this interaction ``task communication''. During task communication the robot must infer the details of the task from unstructured human signals and it must choose actions that facilitate this inference.
In this paper we propose the use of a partially observable Markov decision process (POMDP) for representing the task communication problem;
with the unobservable task details and unobservable intentions of the human teacher captured in the state, with all signals from the human represented as observations, and with the cost function chosen to penalize uncertainty.
We work through an example representation of task communication as a POMDP, and present results from a user experiment on an interactive virtual robot, compared with a human controlled virtual robot, for a task involving a single object movement and binary approval input from the teacher.
~The results suggest that the proposed POMDP representation produces robots that are robust to teacher error, that can accurately infer task details, and that are perceived to be intelligent.
\end{abstract}

\section{Introduction}

General purpose robots such as Willow Garage's PR2 and Stanford's STAIR robot are capable of performing a wide range of tasks such as folding laundry \cite{Berg:2010}, and unloading the dishwasher \cite{Saxena:2006}. While many of these tasks will come pre-programmed from the factory, we would also like the robots to acquire new tasks from their human owners. For the general population, this demands a simple and robust method of communicating new tasks. Through this paper we hope to promote the use of the partially observable Markov decision processes (POMDP) as a framework for controlling the robot during these task communication phases. The idea is that we represent the unknown task as a set of hidden random variables. Then, if the robot is given appropriate models of the human, it can choose actions that elicit informative responses from the human, allowing it to infer the value of these hidden random variables. We formalize this idea in the {\it Framework} section. This approach makes the robot an active participant in task communication.\footnote{Though related, this is different from an \emph{active learning} problem~\cite{Burr:2010}, since the interaction in task communication is less structured than in the supervised learning setting.}

Note that we distinguish ``task communication'' from ``task execution''. Once a task has been communicated it might then be associated with a trigger for later task execution. This paper deals with communicating the details of a task, not commanding the robot to execute a task; i.e. task communication not task execution.

Many researchers have addressed the problem of task communication.
A common approach is to control the robot during the teaching process, and demonstrate the desired task~\cite{Grollman:2007}\cite{Saunders:2006}\cite{Nicolescu:2003}. The problem for the robot is then to infer the task from examples. Our proposal addresses the more general case in which the robot must actively participated in the communication, choosing actions to facilitate task inference.

In other work, as in ours, the task communication is more hands off, requiring the robot to choose actions during the communication, with much of the work using binary approval feedback as in our experiment below ~\cite{Touretzky:1997}\cite{Blumberg:2002}\cite{Kaplan:2002}. Our work differs in that we propose the use of a POMDP representation, while prior work has created custom representations with ad hoc inference and action selection procedures.

An exception is the Sophie's Kitchen work which used the widely accepted MDP representation\cite{Thomaz:2006}. Their approach differs in many ways from ours but most importantly is in the way actions are selected during task communication. In their work the robot repeatedly executes the task, with some noise, as best it currently knows it, while in our approach the robot chooses actions to become more certain about the task. Intuitively, if the goal of the interaction is to communicate a task as quickly as possible, then repeatedly executing the full task as you currently believe it, is likely not the best policy. Instead, the robot should be acting to reduce uncertainty specifically about the details of the task that it is unclear on.
In order to generate these uncertainty reducing actions we feel that a representation allowing for hidden state is needed, and we propose the POMDP. In a POMDP there can be a distribution over details of the task, and actions can be generated to reduce the uncertainty in this distribution.

Substantial work has also been done on human assisted learning of low level control policies, such as the mountain car experiments, where the car must learn a throttle policy for getting out of a ravine~\cite{Knox:2009}. While the mode of input is the same as is used in our experiment (a simple rewarding input signal), we are addressing different problems and different solutions are appropriate. They are addressing the problem of transferring a control policy from a human to the robot, where explicit conversation actions to reduce uncertainty would be inefficient and treating the human input as part of an environmental reward is appropriate. In contrast we are addressing the problem of communicating higher level tasks, such as setting the table, in which case explicitly modeling the human and taking communication actions to reduce uncertainty is beneficial, and treating the human input as observations carrying information about the task details is appropriate.

In the next section we present our approach of representing task communication as a POMDP, and in the following section we describe a user-experiment which demonstrates our approach and highlights the advantages.

\section{Framework}

\subsection{Definitions}

Random variables will be designated with a capital letter or a capitalized word; e.g. $X$. The values of the random variable will be written in lowercase; e.g. $X=x$. If the value of a random variable can be directly observed then we will call it an ``observable'' random variable. If the value of a random variable cannot be directly observed then we will call it a ``hidden'' random variable, also known as a ``latent'' random variable. If the random variable is ``sequential'', meaning it changes with time, then we provide a subscript to refer to the time index; e.g. $M_t$ below. A random variable can be multidimensional; e.g. the state $S$ below is made up of other random variables: $Mov$, $M_t$, etc. If a set of random variables contains at least one hidden random variable then we will call it ``partially observable''.

$P(X)$ is the probability distribution defined over the domain of the random variable $X$. $P(x)$ is the value of this distribution for the assignment of $x$ to $X$. $P(X|Y)$ is a ``conditional'' probability distribution, and defines the probability of a value of $X$ given a value of $Y$. A ``marginal'' probability distribution is a probability distribution that results from summing or integrating out other random variables; e.g. $P(X,Y): \forall_{x \in X, y \in Y}P(x,y)=\sum_{z \in Z}P(x,y,z)$. When a probability distribution has a specific name, such as the ``transition model'' or the ``observation model'' we will use associated letters for the probability distribution; e.g. $T(...)$ or $\Omega(...)$.

\subsection{POMDP Review}

A POMDP representation is applicable to action selection problems where the environment is sequential and partially observable. A POMDP is a 6 element tuple: $\tuple{S,\linebreak[1] A,\linebreak[1] O,\linebreak[1] T,\linebreak[1] \Omega,\linebreak[1] C,\linebreak[1] \gamma}$. $S$ is the set of possible world states; $A$ is the set of possible actions; $O$ is the set of possible observations that the agent may measure; $T$ is the transition model defining the stochastic evolution of the world (the probability of reaching state $s' \in S$ given that the action $a \in A$ was taken in state $s \in S$); $\Omega$ is the observation model that gives the probability of measuring an observation $o \in O$ given that the world is in a state $s \in S$; $C$ is a cost function which evaluates the penalty of a state $s \in S$ or the penalty of a probability distribution over states $b$, $b: \forall_{s \in S} \left[0 \leq b(s) \leq 1 \textnormal{ \emph{and} } \sum_{s \in S} b(s) = 1\right]$; and, finally, $\gamma$ is the discount rate for the cost function. 

Given a POMDP representation, a POMDP solver seeks a policy $\pi$, $\pi(b): b \to a$, that minimizes the sum of discounted expected cost.\footnote{Often a reward function is used instead of a cost function, but these are interchangeable; minimizing the cost function $C$ is the same as maximizing the reward function -$C$.}\footnote{We will use the formulation of $C$ as a cost function over a probability distribution, since in communication we care that the communication is successful, not what was communicated.}
The cost is given by $C$, the discounting is given by $\gamma$, and the expectation is computed using $T$ and $\Omega$.

For further reading on POMDPs see ~\cite{Sondik:1971}, ~\cite{Thrun:2005}, and ~\cite{Kaelbling:1998}. For an introduction to POMDP solvers, with further references, see ~\cite{Thrun:2005} chapter 15. 

POMDPs have been successfully applied to problems as varied as autonomous helicopter flight~\cite{Ng:2003}, mobile robot navigation~\cite{Roy:2002}, and action selection in nursing robots~\cite{Pineau:2003}.

\subsection{Approach}

We propose the use of a POMDP for representing the problem of a human communicating a task to a robot. Specifically, for the elements of the POMDP tuple $\tuple{S,\linebreak[1] A,\linebreak[1] O,\linebreak[1] T,\linebreak[1] \Omega,\linebreak[1] C,\linebreak[1] \gamma}$, the partially observable state $S$ captures the details of the task along with potentially the mental state of the human (helpful for interpreting ambiguous observations); the set of actions $A$ capture all possible actions of the robot during communication (for example words, gestures, body positions, etc.); the set of observations $O$ capture all signals from the human (be they words, gestures, buttons, etc.); and the cost function $C$ should encode the desire to minimize uncertainty over the task details in $S$.
The transition model $T$, the observation model $\Omega$, and the discount rate $\gamma$ fill their usual POMDP roles. The experiment below provides an example of $T$ and $\Omega$ for task communication.

\section{Experiment}

\subsection{Simulator}

\begin{figure}[t]
  \centering
  \includegraphics{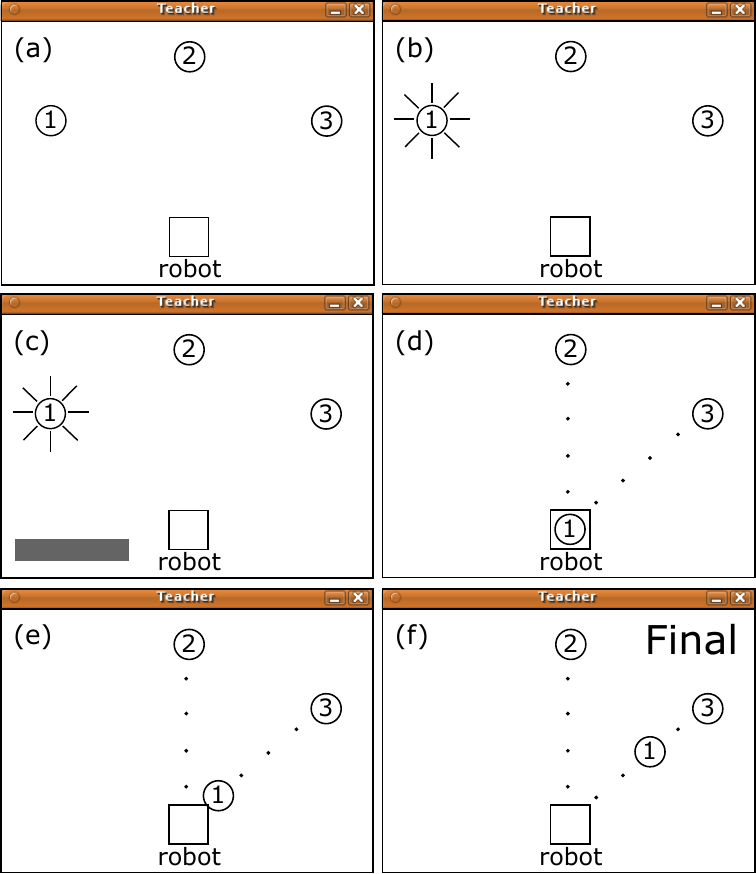}
  \caption{Typical human-robot interactions on the simulator. (a) The state all trials start in; the square robot, holding no balls, surrounded by the three balls. (b) The robot lighting one of the balls, as if to ask, ``move this?''. (c) The human teacher pressing the keyboard spacebar, displayed in the simulator as a rectangle, and used to indicate approval with something the robot has done. (d) The robot holding one of the balls, the four relative distances are displayed to the remaining two balls. (e) The robot has slid one of the balls to the furthest distance from another ball. (f) the robot has displayed ``Final'', indicating that it knows the task, and that the world is currently in a state consistent with that task.}
  \label{fig:virtual_world}
\end{figure}

The experiment was run using the simulator shown in figure \ref{fig:virtual_world}. The virtual world consists of 3 balls and the robot, displayed as circles and a square (\ref{fig:virtual_world}.a). The robot can ``gesture'' at balls by lighting them (\ref{fig:virtual_world}.b), it can pick up a ball (\ref{fig:virtual_world}.d), it can slide one ball to one of four distances from another ball (\ref{fig:virtual_world}.e), and it can signal that it knows the task by displaying ``Final'' (\ref{fig:virtual_world}.f). For the comparison trials, in which a human is controlling the robot, these action are controlled by left and right mouse licks on the objects involved in the action; e.g. right click on a ball to light it or turn off the light, left click on a ball to pick it up, etc. In this interface the teacher input is highly constrained; the teacher has only one control and that is the keyboard spacebar, visually indicated by a one half second rectangle (\ref{fig:virtual_world}.c), and is used to indicate approval with something that the robot is doing or has done. The simulator is networked so that two views can be opened at once; this is important for the comparison trials, where the human controlling the robot must be hidden from view.

Timesteps are 0.5 second long, i.e. the robot receives an observation and must generate an action every 0.5 seconds. The simulator is free running, so, in the comparison trials where the human controls the robot, if the human does not select an action, then the ``no action'' action is taken. ``no action'' actions are still taken in the case of the POMDP controlled robot, but they are always intentional actions that have been selected by the robot. We provide enough processing power so that the POMDP controlled robot always has an action ready in the allotted 0.5 seconds. 

The simulated world is discrete, observable, and deterministic.

\subsection{Toy Problem}

For this experiment, the human teacher was asked to communicate, through only spacebar presses, that a specific ball should be at a specific distance from another specific ball. The teacher was told that a spacebar press would be interpreted by the robot as approval of something that it was doing. At the beginning of each trial the teacher was shown a card designating the ball relationship to teach. The robot, either POMDP or human controlled, had to infer the relationship from spacebar presses (elicited from its actions).
When the robot thought it knew the relationship, it would end the trial by moving the world to that relationship and displaying ``Final''.
The teacher would then indicate on paper whether the robot was correct and how intelligent they felt the robot in that trial was.

Figure \ref{fig:virtual_world} shows snap shots from one typical trial. The robot ``questioned'' which ball to move (\ref{fig:virtual_world}.b), the teacher indicated approval (\ref{fig:virtual_world}.c), the robot picked up the ball (\ref{fig:virtual_world}.d), the robot generally ``questioned'' which ball to move the one it was holding to (not shown), the robot slid the ball toward another ball (\ref{fig:virtual_world}.e), the teacher approved a distance or progress toward a distance (not shown), and, after further exploration, the robot indicated that it knew the task by displaying ``Final'' (\ref{fig:virtual_world}.f).

Although this problem is simplistic, 
a robot whose behaviors consist of chainings of these simple two object relationship tasks could be useful; e.g. for the ``set the table'' task: move the plate to zero inches from the placemat, move the fork to one inch from the plate, move the spoon to one inch from the fork, etc.

We chose the spacebar press as the input signal in our experiment because it carries very little information, requiring the robot to infer meaning from context, which is a strength of our approach. For a production robot, this constrained interface should likely be relaxed to include signals such as speech, gestures, or body language. These other signals are also ambiguous, but the simplicity of a spacebar press made the uncertainty obvious for our demonstration.

\subsection{Formulation}

\subsubsection{State ($S$)}

The state $S$ is composed of hidden and observable random variables.

The task that the human wishes to communicate is captured in three hidden random variables $Mov$, $WRT$, and $Dist$. $Mov$ is the index of the ball to move ($1-3$). $WRT$ is the index of the ball to move ball $Mov$ with respect to. $Dist$ is the distance that ball $Mov$ should be from ball $WRT$.

The state also includes a sequential hidden random variable, $M_t$, for interpreting the observations $O_t$. $M_t$ takes on one of five values: ($waiting$, $mistake$, $that\_mov$, $that\_wrt$, or $that\_dist$). A value of $waiting$ implies that the human is waiting for some reason to press the spacebar. A value of $mistake$ implies that the human accidentally pressed the spacebar. A value of $that\_mov$ implies that the human pressed the spacebar to indicate approval of the ball to move. A value of $that\_wrt$ implies that the human pressed the spacebar to indicate approval of the ball to move ball $Mov$ with respect to. A value of $that\_dist$ implies that the human pressed the spacebar to indicate approval of the distance that ball $Mov$ should be from ball $WRT$.

The state also includes observable random variables for the physical state of the world; e.g. which ball is lit, which ball is being held, etc.

Finally, the state includes ``memory'' random variables for capturing historical information, e.g. the last time step that each of the balls were lit, or the last time step that $M = that\_mov$. The historical information is important for the transition model $T$. For example, humans typically wait one to five seconds before pressing the spacebar a second time. In order to model this accurately we need the time step of the last spacebar press.

\begin{center}
\begin{tabular}{|r|l|}
  \hline
  state & Mov \\
  & WRT \\
  & Dist \\
  & M \\
  & $\langle$world state variables$\rangle$ \\
  & $\langle$historical variables$\rangle$ \\
  \hline
\end{tabular}
\end{center}

\subsubsection{Actions ($A$)}

There are six parameterized actions that the robot may perform. Certain actions may be invalid depending on the state of the world. The actions are: $noa$, for performing no action and leaving the world in the current state; $light\_on(index)$ or $light\_off(index)$, for turning the light on or off for the ball indicated by $index$; $pick\_up(index)$, for picking up the ball indicated by index; $release(index)$, for putting down the ball indicated by index; and $slide(index\_1,\linebreak[1] distance,\linebreak[1] index\_2)$, for sliding the ball indicated by $index\_1$ to the distance indicated by $distance$ relative to the ball indicated by $index\_2$. Note that only a few actions are valid in any world state; for example, $slide(index\_1,\linebreak[1] distance,\linebreak[1] index\_2)$ is only valid if ball $index\_1$ is currently held or currently at a distance from ball $index\_2$ and if $distance$ is only one step away from the current distance.

\begin{center}
\begin{tabular}{|r|l|}
  \hline
  actions & noa \\
  & light\_on(index) \\
  & light\_of(index) \\
  & pick\_up(index) \\
  & release(index) \\
  & slide(index\_1, distance, index\_2) \\
  \hline
\end{tabular}
\end{center}

\subsubsection{Observations ($O$)}

An observation takes place at each time step and there are two valid observations: $spacebar$ or $no\_spacebar$, corresponding to whether the human pressed the spacebar on that time step.

\begin{center}
\begin{tabular}{|r|l|}
  \hline
  observations & spacebar \\
  & no\_spacebar \\
  \hline
\end{tabular}
\end{center}

\subsubsection{Transition Model ($T$)}

A transition model gives the probability of reaching a new state, given an old state and an action. In this example, $Mov$, $WRT$, and $Dist$ are non-sequential random variables, meaning they do not change with time, so $T(Mov=i,...|Mov=i,...)=1.0$. The transition model for the physical state of the virtual world is also trivial, since the virtual world is deterministic.

\begin{figure}[t]
  \centering
  \includegraphics[scale=1.0]{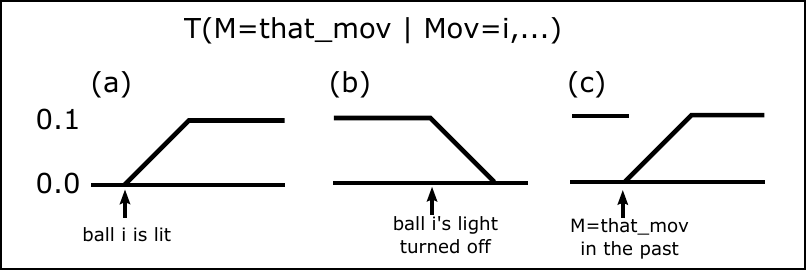}
  \caption{This is an illustration of part of the transition model $T$. Here we show the probability that the human will signal their approval (via a spacebar press) of the ball to be moved, $T(M=that\_mov|Mov=i,...)$, where, in this hypothesis, the ball to be moved is ball $i$. (a) once the robot has lit ball $i$, the probability increases from zero to a peak of 0.1 over 2 seconds. (b) after the light has turned off there is still probability of an approving spacebar press, but decreasing over 2 seconds. (c) If the teacher has signaled their approval ($M=that\_mov$), then the probability resets. The structure and shape of these models was set from intuition.}
  \label{fig:table_top}
\end{figure}

The variable of interest in this example for the transition model is the sequential random variable $M$ that captures the mental state of the human ($waiting$, $mistake$, $that\_mov$, $that\_wrt$, or $that\_dist$). The transition model was specified from intuition, but in practice we envision that it would either be specified by psychological experts, or learned from human-human or human-robot observations. For the experiment we set the probability that $M$ transitions to $mistake$ from any state to a fixed value of 0.005, meaning that at any time there is a 0.5\% chance that the human will \emph{mistakenly} press the spacebar indicating approval. We define the probability that $M$ transitions to $that\_mov$, $that\_wrt$, or $that\_dist$ as a table-top function, as shown in figure:~\ref{fig:table_top}.
We set the probability that $M$ transitions to $waiting$ to the remaining probability; $T(M=waiting) = 1-T(M=mistake \vee that\_mov \vee that\_wrt \vee that\_dist).$

\subsubsection{Observation Model ($\Omega$)}

The observation model we have chosen for this problem is a many to one deterministic mapping:
\[
P(O=spacebar|M) = 
\left\{ 
  \begin{array}{ll}
    0.0 & \mbox{if $M = waiting$} \\
    1.0 & \mbox{otherwise}
  \end{array}
\right.
\]

Note that this deterministic model does not imply that the state is observable since, given $O = spacebar$, we do not know why the human pressed the spacebar, $M \stackrel{?}{=} (mistake \vee that\_mov \vee that\_wrt \vee that\_dist)$.
\footnote{If there were noise in the spacebar key then this would not be a deterministic mapping.}

\subsubsection{Cost Function ($C$)}

As mentioned earlier, the cost function should be chosen to motivate the robot to quickly and accurately infer what the human is trying to communicate. In our case this is a task captured in the random variables $Mov$, $WRT$, and $Dist$. The cost function we have chosen is the entropy of the marginal distribution over $Mov$, $WRT$, and $DIST$:\footnote{By ``entropy'' we mean information entropy, not thermodynamics entropy.}
\begin{equation}
C(p) = -\sum_x p(x) \cdot log(p(x)).
\end{equation}
Where $p$ is the marginal probability distribution over $Mov$, $WRT$, and $Dist$, and $x$ takes on all permutations of the value assignments to $Mov$, $WRT$, and $Dist$.

Since entropy is a measure of the uncertainty in a probability distribution, this cost function will motivate the robot to reduce its uncertainty over $Mov$, $WRT$, and $Dist$, which is what we want.

\subsubsection{Discount Rate ($\gamma$)}

$\gamma$ is set to 1.0 in our experiment, meaning that uncertainty later is just as bad as uncertainty now. The valid range of $\gamma$ for a POMDP solver evaluating actions to an infinite horizon is $0 \leq \gamma < 1.0$, but our solver only evaluates to a 2.5 second horizon. In practice, the larger the value of $\gamma$, the more willing the robot is to defer smaller gains now for larger gains later.

\subsection{Action Selection}

The problem of action selection is the problem of solving the POMDP. 
There are many established techniques for solving POMDPs~\cite{Murphy:2000}. In our case, given the simplicity of the world and the problem, we could take a direct approach. The robot expands the action-observation tree out 2.5 seconds into the future, and takes the action that minimizes the sum of expected entropy over this tree.
This solution is approximate, since we only look ahead 2.5 seconds, but, as we will show shortly, it results in reasonable action selections for the toy problem used in the experiment.

When the marginal probability of one of the assignments to $Mov$, $WRT$, and $Dist$ is greater than 0.98 (over 98\% confident in that assignment), then the robot moves the world to that assignment and displays ``Final''.
\subsection{Subjects}

The experiment involved 26 participants, consisting of undergraduate and graduate students ranging in age from 18 to 31 with a mean age of 22. Four of the participants were randomly selected for the ``human robot'' role, leaving 22 participants for the ``teacher'' role. The participants rated their familiarity with artificial intelligence software and systems on a scale from 1 to 7; the mean score was 3.4 with a standard deviation of 1.9. Participants were paid \$10.00 for their time in the experiment.

\subsection{Results}

\subsubsection{Is Robust to Mistakes}

\begin{figure}[t]
  \includegraphics[width=3.2in]{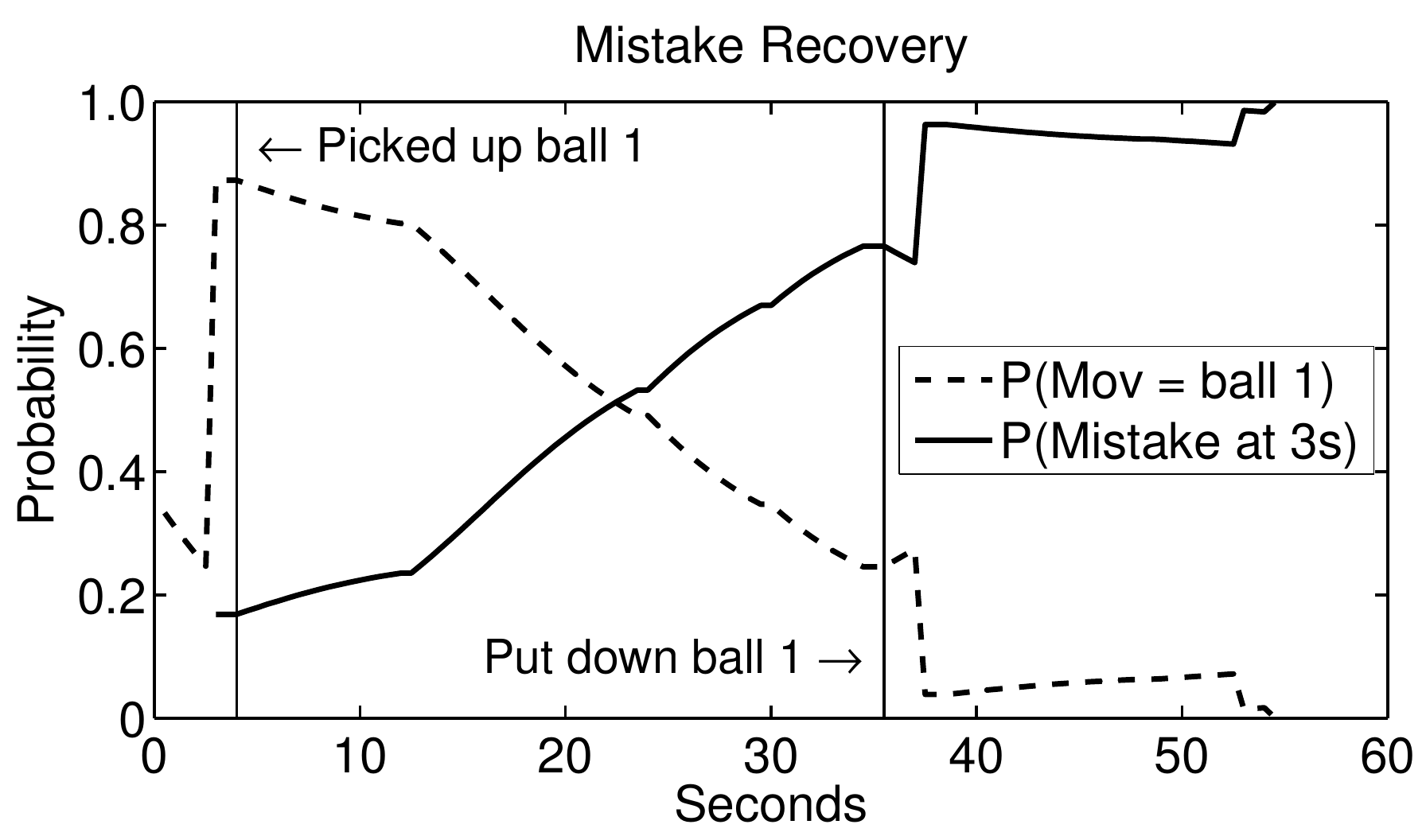}
  \caption{This figure shows a typical recovery from a mistaken spacebar press. In this figure the teacher mistakenly pressed the spacebar at the three second mark while the robot was lighting ball 1. The probability that ball 1 was the ball to be moved immediately spiked. At the same time there was a low probability that the spacebar press was a mistake. At 4 seconds the robot picked up ball 1 and started moving it, exploring tasks involving the movement of ball 1. As the trial progressed without further spacebar presses, the probability that the spacebar press at 3 seconds was a mistake increased and the probability that ball 1 was the ball to move decreased. Finally, at 36 seconds the approximately optimal policy was to put down ball 1 and reassess which object was to be moved.}
  \label{fig:mistake_recovery}
\end{figure}

The strength of using a probabilistic approach such as a POMDP is in its robustness to noise. In our experiment noise came in the form of mistaken spacebar presses. Figure~\ref{fig:mistake_recovery} illustrates a typical mistaken spacebar press. In this trial, at the three second mark, the human mistakenly pressed the spacebar while ball 1 was lit, when in fact ball 1 was not involved in the task. As expected, the robot's marginal probability that ball 1 was the ball to move immediately spiked. Yet there was still a small probability that the random variable $M$ equaled $mistake$ at the three second mark. The trial proceeded with the robot making use of the strong belief that ball 1 was the ball to be moved: it picked up ball 1 at 4 seconds and lit ball 2 and ball 3. As time progressed, and the robot did not receive further spacebar presses that would be consistent with a task involving ball 1, the probability that the human mistakenly pressed the spacebar increased and the probability that ball 1 was the ball to move decreased. At thirty six seconds, the belief that a mistake occurred was strong enough that the action which minimized the expected entropy was to put down ball 1 and continue seeking another ball to move.

\subsubsection{Can Infer Hidden State}

\begin{figure}[t!]
  \centering
  \subfigure[Ball to mov] {
    \includegraphics[scale=0.24]{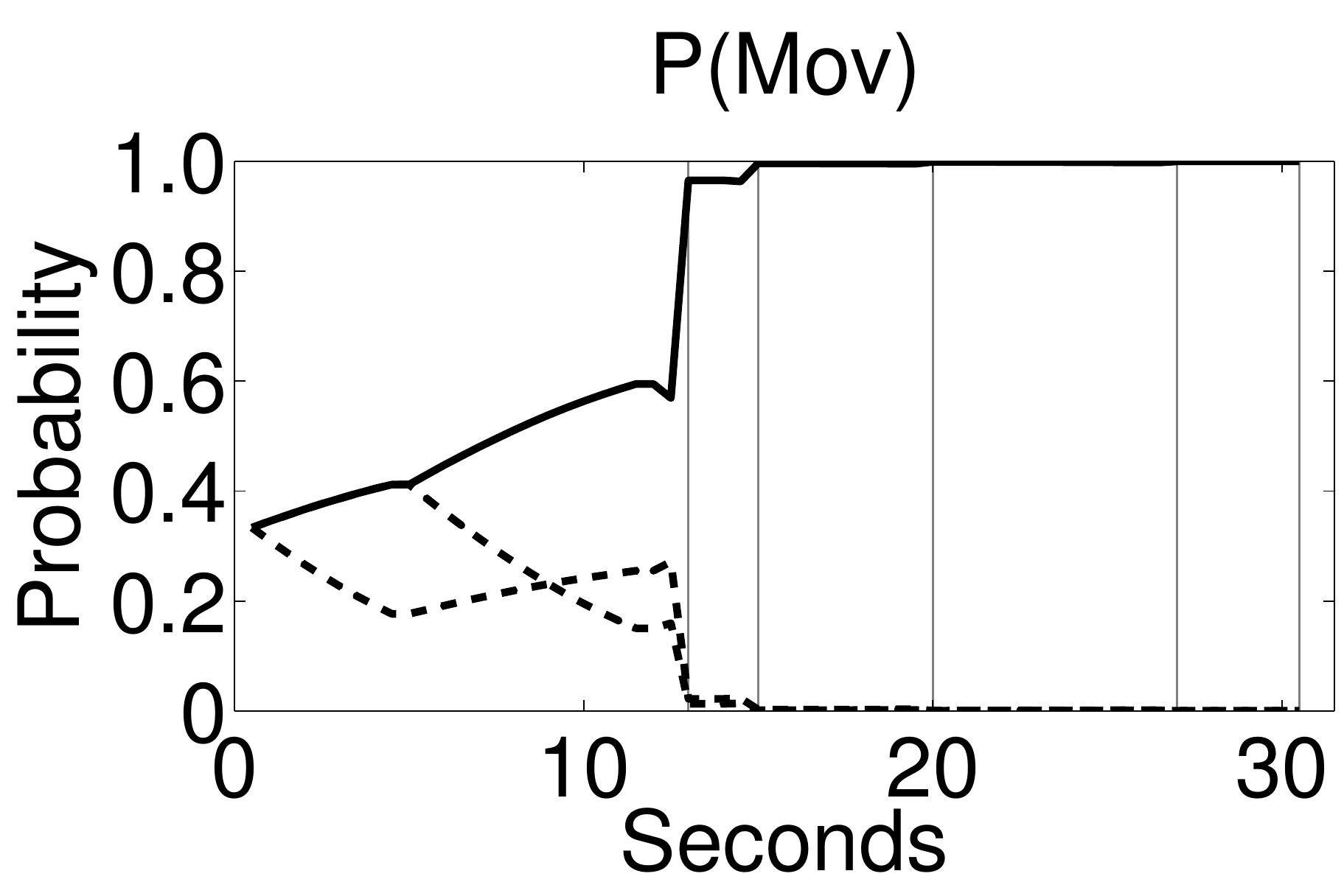} 
    \label{fig:task_marginals_ob}
  }
  \subfigure[Ball to mov with respect to] {
    \includegraphics[scale=0.24]{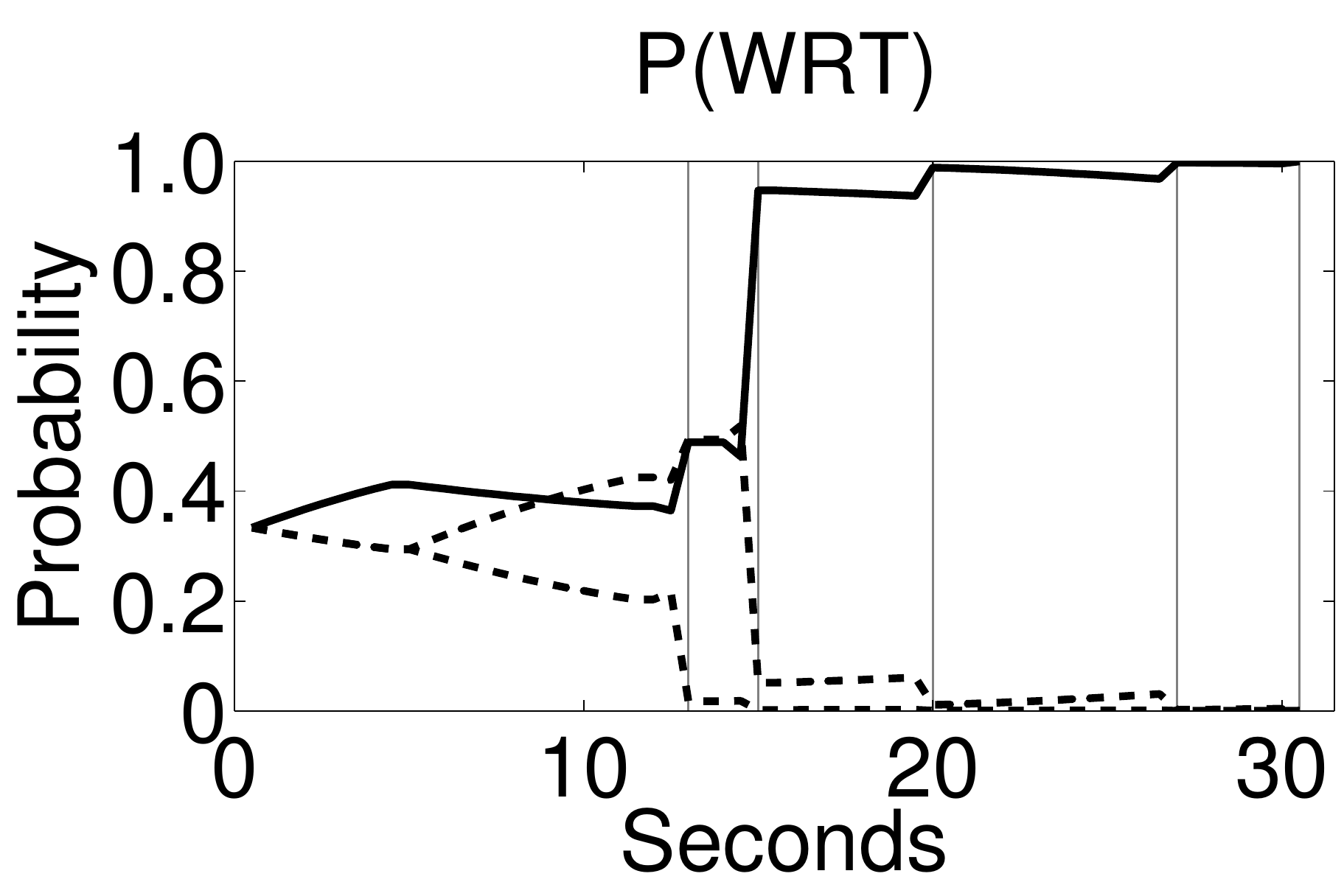}
    \label{fig:task_marginals_wrt}
  }
  \subfigure[Distance from ball WRT] {
    \includegraphics[scale=0.24]{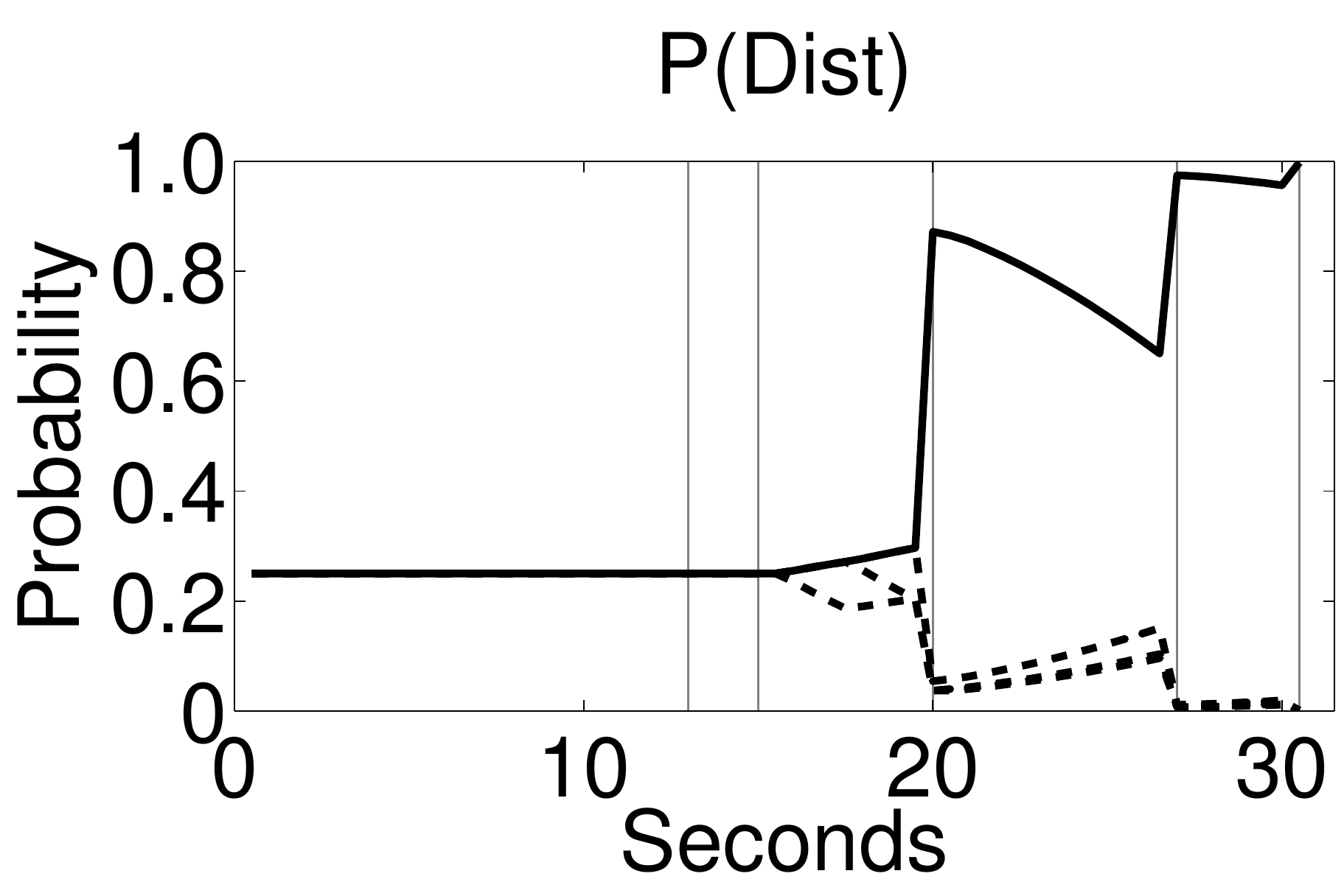}
    \label{fig:task_marginals_dist}
  }
  \caption{This figure shows the robot's inference of the ball to mov (a), the ball to move it with respect to (b), and the distance between the balls (c) for one of the trials. The vertical lines designate spacebar presses. The solid line in each figure shows the marginal probability of the true assignment for that random variable. The marginal probabilities for the true assignments are driven to near 1.0 by information gathered from the spacebar presses elicited by the robot's actions.}
  \label{fig:task_marginals}
\end{figure}

\begin{figure}[t]
  \centering
  \includegraphics[width=3.2in]{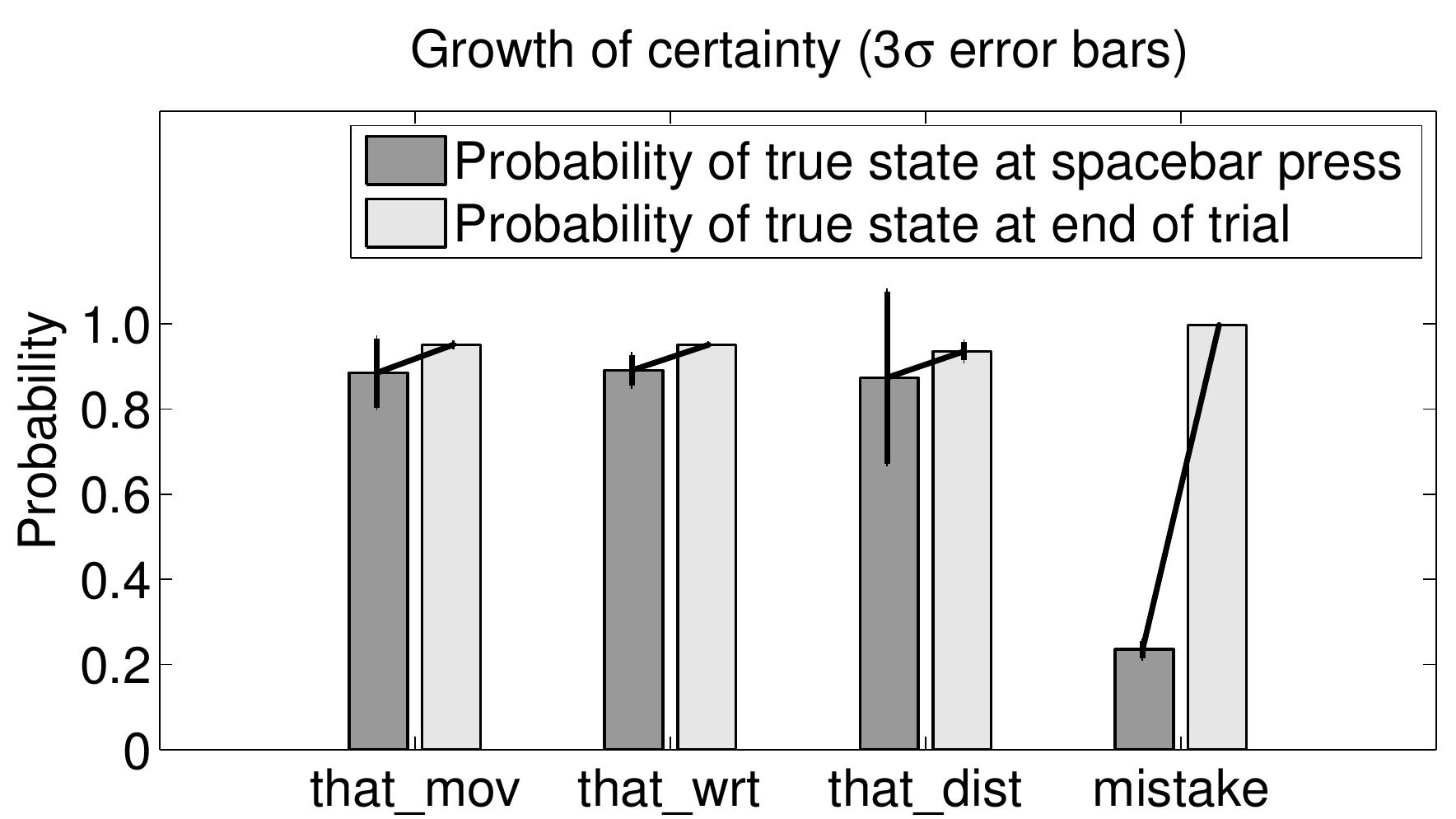}
  \caption{This figure shows the marginal probability of the four approval mental states at the time the spacebar was pressed (dark gray) and at the end of the trial (light gray). The true states were labeled in a post processing step. All spacebar presses from all 22 POMDP trials are included. This shows that, for each of the mental states, the marginal probability of the correct state increases as the trial progresses and ends at near certainty. This is most pronounced in the case of $M=mistake$, in which the initial probability that the spacebar was a mistake is low, but increases dramatically as the trial progresses.}
  \label{fig:accuracy_of_inference}
\end{figure}

The second result from the experiment is that the robot accurately inferred the hidden task and the hidden state of the teacher. In all trials the human teachers reported that the robot was correct about the task being communicated. Figure~\ref{fig:task_marginals} shows a look at the robot's marginal probabilities, for one of the trials, of the random variables $Mov$, $WRT$, and $Dist$. 
In this trial, as was typical of the trials, the robot first grew its certainty about $Mov$ followed by $WRT$ and then $Dist$. Figure~\ref{fig:accuracy_of_inference} shows the probability of the true assignment to $M$ at the time of the spacebar press and at the end of the trial, for four assignments to the variable $M$.\footnote{We did not include \emph{before} and \emph{after} for $M=waiting$ because the observation model $\Omega$ makes this assignment deterministic.} 
This shows that as each trial progressed the robot became correctly certain about what the human meant by each spacebar press.

\subsubsection{Can Reasonably Select Actions}

We captured two metrics in an effort to evaluate the quality of the POMDP selected actions.

\begin{figure}[h]
  \includegraphics[width=3.2in]{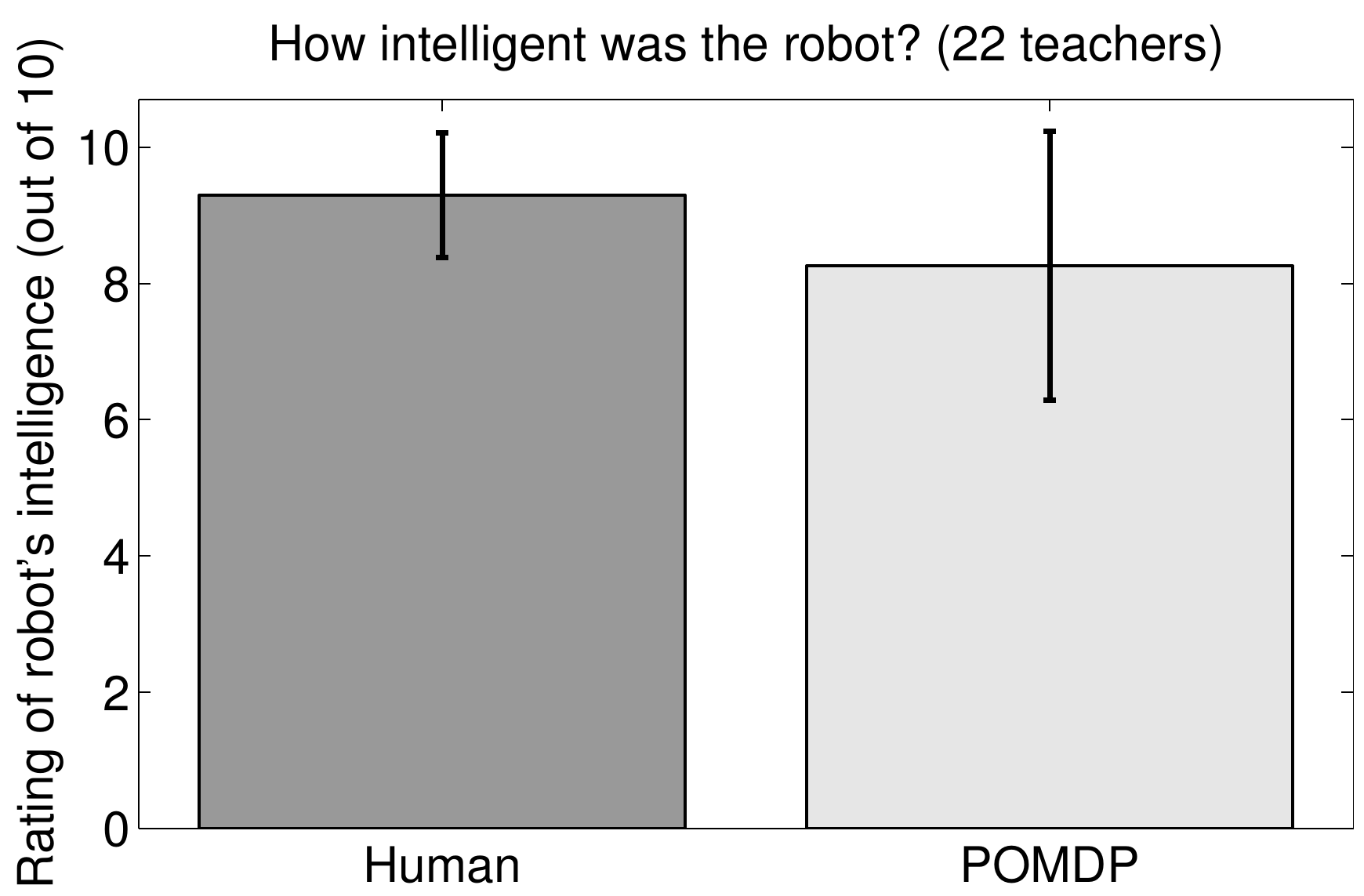}
  \caption{The 22 human teachers each participated in two trials, one teaching the human controlled robot and one teaching the POMDP controlled robot. The order of human or POMDP robot was randomized, and the true identity of the robot controller was hidden from the teacher. Following each trial the teacher rated the intelligence of the robot on a scale from 1 to 10, with 10 being the most intelligent. With the exception of one teacher, all teachers rated the human controlled robot the same or more intelligent than the POMDP controlled robot (mean of 9.30 vs. 8.26).}
  \label{fig:intelligent}
\end{figure}

The first was a subjective rating of the robot's intelligence by the teacher after each trial. Figure~\ref{fig:intelligent} shows the ratings for the human controlled robot and the ratings for the POMDP controlled robot. The human received higher intelligence ratings, but not significantly; we believe that this gap can be improved with better modeling (see section~\ref{sec:improve_models}).

\begin{figure}[h]
  \includegraphics[width=3.2in]{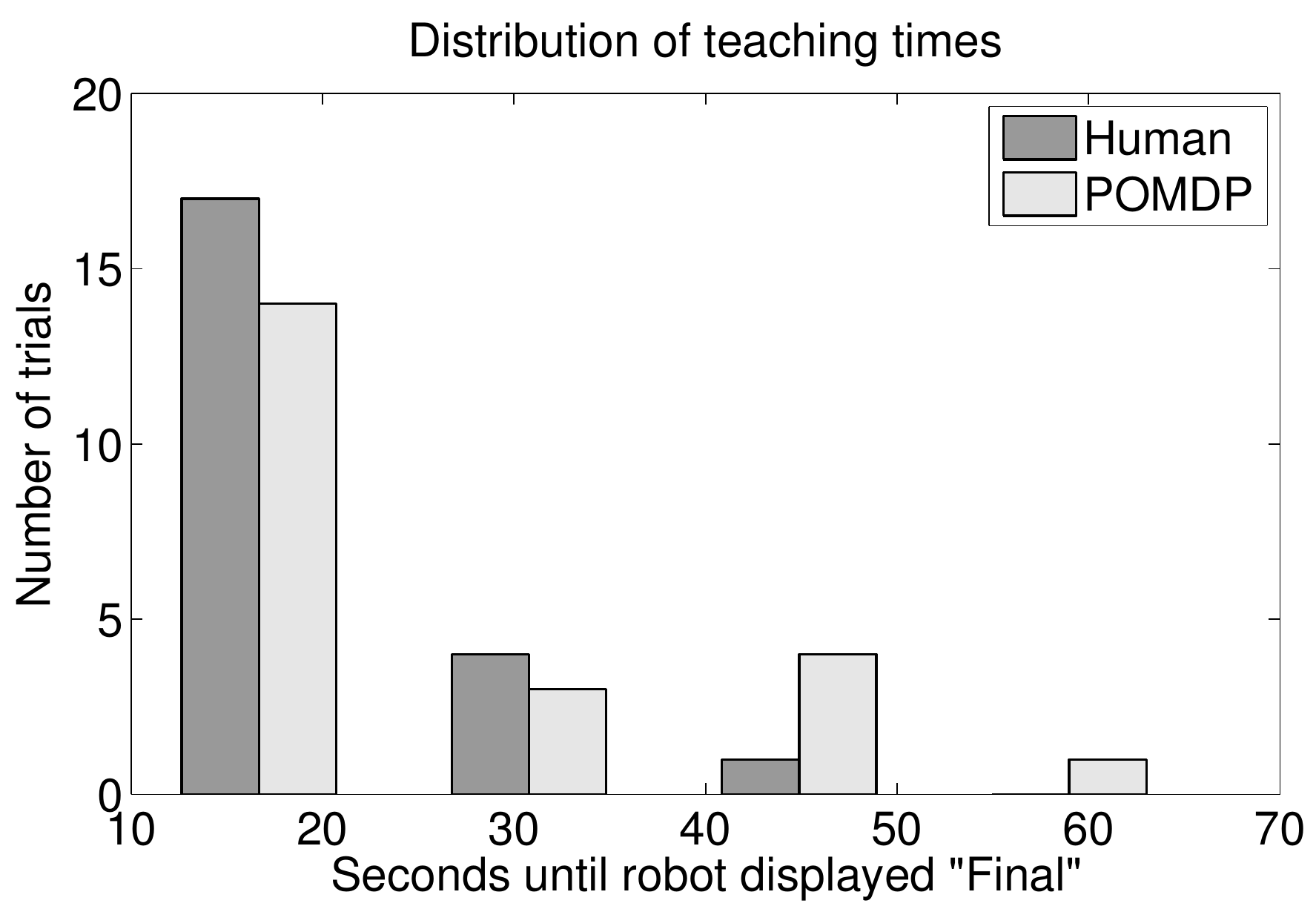}
  \caption{A histogram of the times until the robot, human or POMDP controlled, displayed ``Final''. The robot displayed ``Final'' to signal that they knew the task and that the world was displaying the task. The POMDP controlled robot displayed ``Final'' when the marginal probability for a particular task, $P(Mov=i,WRT=j,Dist=k)$, was greater than 0.98. In all trials the robot, human or POMDP controlled, correctly inferred the task. Task communication, as expected, took longer for the POMDP controlled robot than for the human controlled robot.}
  \label{fig:time_to_teach}
\end{figure}

The second metric we looked at was the time to communicate the task. This was measured as the time until the robot displayed ``Final''. Figure~\ref{fig:time_to_teach} is a histogram of the time until the robot displayed ``Final'' for the POMDP robot and for the human controlled robot. Here again the human controlled robot outperformed the POMDP controlled robot, but the POMDP controlled robot performed reasonably well. Part of this discrepancy could be due to inaccurate models, as in the intelligence ratings, but in this case we believe that the threshold for displaying ``Final'' was higher for the POMDP robot (over 98\% confident) than for the human. Notably, we often observed the human controlled robot displaying ``Final'' after a single spacebar press at the final location. In contrast, the POMDP robot always explored other distances; presumably to rule out the possibility that the first spacebar press was a mistake. Only after a second spacebar press would the POMDP robot display ``Final''.

\begin{figure}[h]
  \includegraphics[width=3.2in]{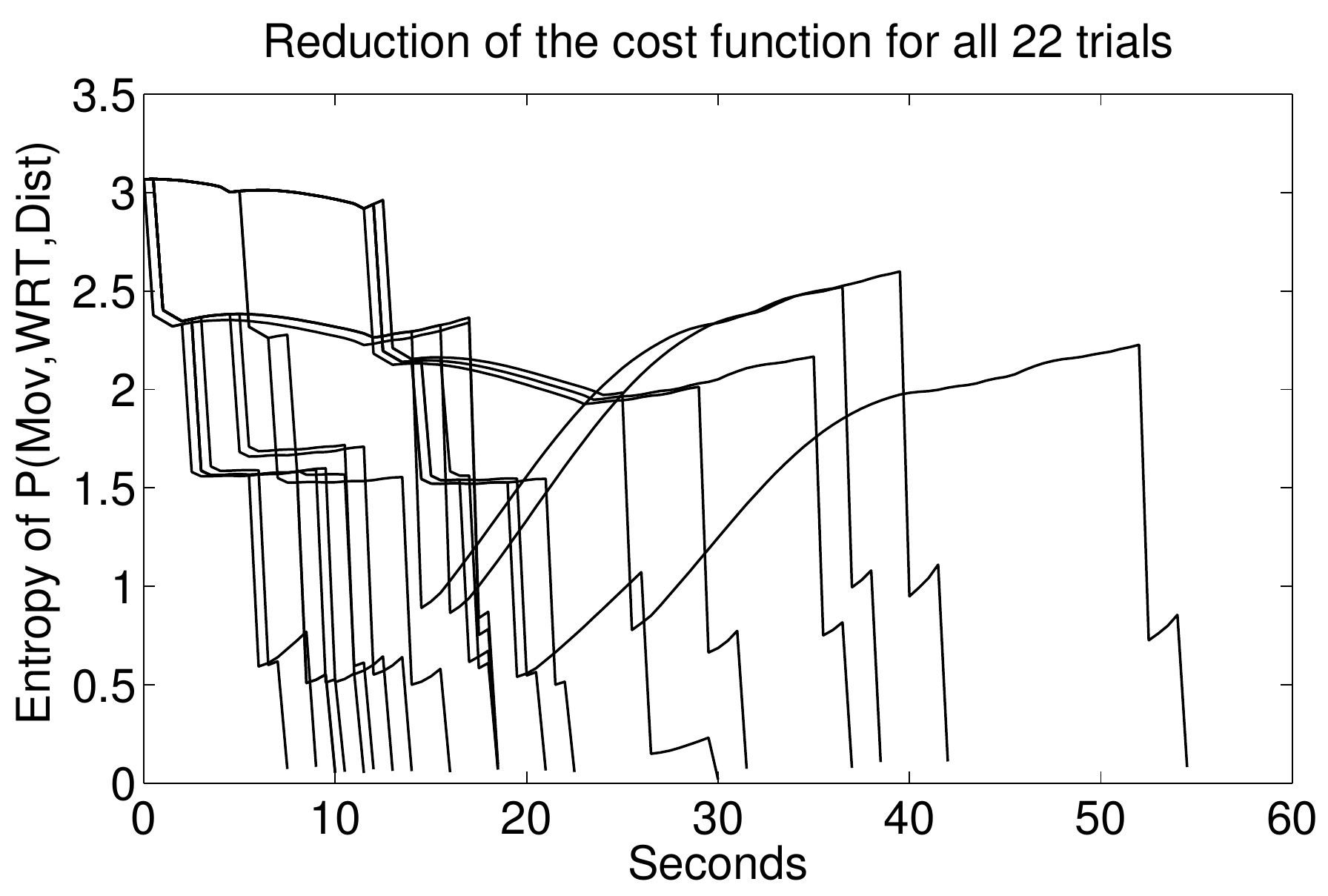}
  \caption{This figure shows the decrease in the cost function over time for all 22 trials of the POMDP controlled robot. The cost function used was the entropy of the marginal distribution over the $Mov$, $WRT$, and $Dist$ random variables. All trials began at the same entropy, the entropy of the uniform distribution over $Mov$, $WRT$, and $Dist$. In all trials the entropy was driven to near zero in less than one minute. Rapid drops correlate with spacebar presses, while large increases correspond to trials where the teacher mistakenly pressed the spacebar.}
  \label{fig:entropy}
\end{figure}

Of interest as well is the POMDP robot's ability to drive down the cost function over each trial. Figure~\ref{fig:entropy} plots the cost function (entropy) as a function of time for each of the trials. During several trials the entropy increased significantly before dropping again. This corresponds to the trials in which the teacher mistakenly pressed the spacebar; the POMDP robot initially believed that there was information in the key press, but over time realized that it was a mistake and carried no information. The figure shows the reduction of entropy in all trials to near zero.

\section{Future Work}

\subsection{Learning Model Structure and Model Parameters}
\label{sec:improve_models}

In our experiment the structure and parameters of $T$ and $\Omega$ were set from intuition. We believe that both the structure and the parameters of the models can be ``learned'', in the machine learning sense. The models could be learned either from observations of humans communicating with other humans or from observations of humans communicating with robots. In work presently under development, we are fitting the parameters of our model to logs of human teachers teaching human controlled robots. We intend to continue along this line, as it may be unrealistic to expect social scientists to accurately and exhaustively model human teachers. 

\subsection{Complex Tasks}

In our experiment the task communicated consisted of a single object movement. Future work will aim to communicate chains of object movements.

\subsection{Complex Signals}

In our experiment we limited observations to the spacebar key presses. As we move to tasks involving object movements in the real world we plan to incorporate further observations such as the gaze direction of the teacher and pointing gestures from the teacher, perhaps using a laser pointer ~\cite{Kemp:2008}. Note that social behavior such as shared attention, argued for in~\cite{Scassellati:1999}, where the robot looks at the human to see where they are looking, would naturally emerge once the teacher's gaze direction, along with the appropriate models, is added as an observation to the system; knowing what object the human is looking at is informative (reduces entropy), so actions leading to the observation of the gaze direction would have low expected entropy and would likely be chosen.

\subsection{Processing}

Substantial progress has been made towards efficient solutions of POMDPs \cite{Murphy:2000}, yet processing remains a significant problem for POMDPs with complex domains. In future work we hope to improve on these techniques by leveraging parallel processing resources such as server clusters (e.g. Amazon's Elastic Compute Cluster (EC2)), or graphics cards enabled for general purpose computing (GPGPUs).

\subsection{Smooth Task Communication and Task Execution Transitions}

This paper focused on task communication, but a robot will also spend time executing communicated tasks. We would like for the formulation to apply to the entire operation of the robot; with optimal transitions between task communication and task execution. A choice of a broader cost function will be an important first step.
One choice for this cost function might be the cost to the human under the human's cost function. The human's cost function would be captured in random variables, perhaps through a non-parametric model. The POMDP solver could then choose actions which would inform it about the human's cost function, which would aid in minimizing the cost to the human. Note that task communication would still occur under this cost function; for example, the robot might infer that doing a task is painful to the human, and communication would allow the robot to do this task for the human, thus performing communication actions would be attractive.\footnote{Research has shown that inferring another agent's cost function is possible (see inverse reinforcement learning)\cite{Ng:2000}.}

\subsection{IPOMDP}

In a classical POMDP the world is modeled as stochastic, but not actively rational; e.g. days transition from sunny to cloudy with a certain probability, but not as the result of the actions of an intelligent agent. In a POMDP the agent is the only intelligence in the world. An Interactive POMDP (IPOMDP) is one of several approaches that extend the POMDP to multiple intelligent agents \cite{Gmytrasiewicz:2005}. It differs from game theoretic approaches in that it takes the perspective of an individual agent, rather than analyzing all agents globally; the individual agent knows that there are other intelligent agents in the world acting to minimize some cost function, but the actions of those agents and their cost functions may be only partially observable. We feel that the task communication problem falls into this category. The human teacher has objectives and reasons for communicating the task, knowing those reasons could allow the robot to better serve the human. Understanding the human and their objectives is important to the smooth communication and execution transitions described before. Thus future work will extend our framework from the POMDP representation to the IPOMDP representation.

Unfortunately, an IPOMDP adds exponential branching of inter-agent beliefs to the already exponential branching of probability space and action-observations in a POMDP. Thus, while it is a more accurate representation it does make a hard problem even harder. That said, an IPOMDP may serve as a good formulation that we then seek approximate solutions for.

\addtolength{\textheight}{-1.1in}
\section{Summary}

In this paper we proposed the use of a POMDP for representing the human-robot task communication problem, and we demonstrated the approach through a user experiment. The experiment suggested that this representation results in robots that are robust to teacher error, that can accurately infer task details, and that are perceived to be intelligent. Future work will include more complex tasks and signals, better modeling of the teacher, and improved processing techniques.

\bibliographystyle{IEEEtran}
\bibliography{woodward}

\end{document}